\documentclass[10pt,twocolumn,letterpaper]{article}

\usepackage{cvpr}              % To produce the CAMERA-READY version
\usepackage[dvipsnames]{xcolor}
\usepackage{comment}

\newcommand{\yinyu}[1]{\textcolor{blue}{\emph{Yinyu:~{#1}}}}

\definecolor{cvprblue}{rgb}{0.21,0.49,0.74}
\usepackage[pagebackref,breaklinks,colorlinks,citecolor=cvprblue]{hyperref}
\usepackage{diagbox}
\usepackage{graphicx}
\usepackage[inkscapeformat=png]{svg}

\def\paperID{3598} % *** Enter the Paper ID here
\def\confName{CVPR}
\def\confYear{2024}

\title{LASA: Instance Reconstruction from Real Scans using A Large-scale \\Aligned Shape Annotation Dataset}

\author{Haolin Liu\textsuperscript{\rm 1,2}$^{*}$, Chongjie Ye\textsuperscript{\rm 1,2}$^{*}$, Yinyu Nie\textsuperscript{\rm 3}, Yingfan He\textsuperscript{\rm 1,2}, Xiaoguang Han\textsuperscript{\rm 2,1}$^{\dag}$ \\
\small{$^{*}$equal contribution} \qquad \small{$^{\dag}$corresponding author} \vspace{5pt}\\
\textsuperscript{\rm 1}{FNii, CUHKSZ} \qquad \textsuperscript{\rm 2}{SSE, CUHKSZ} \qquad \textsuperscript{\rm 3}{Technical University of Munich} \vspace{5pt}\\
\small{\href{https://gap-lab-cuhk-sz.github.io/LASA/}{gap-lab-cuhk-sz.github.io/LASA}}
}

\begin{document}
\twocolumn[{%
\renewcommand\twocolumn[1][]{#1}%
\maketitle
\begin{center}
    \captionsetup{type=figure}
    \centering
    \includegraphics[width=\linewidth]{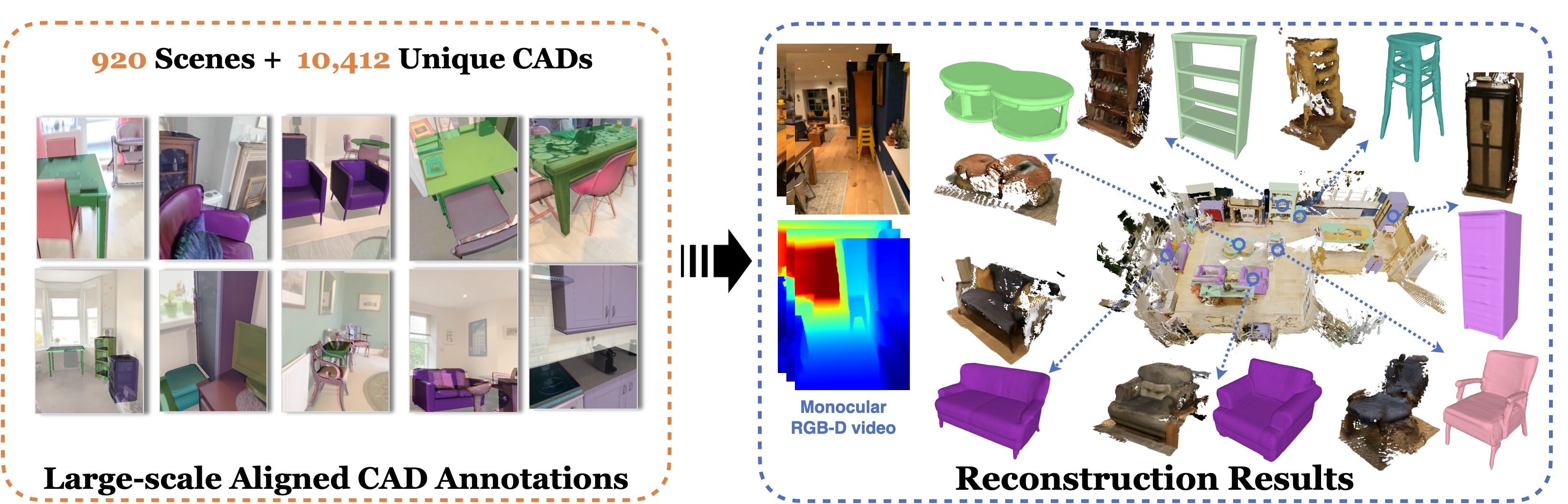}
    \caption{We introduce LASA, a Large-scale Aligned Shape Annotation Dataset containing 10,412 unique object CAD models aligned with 920 real-world scene scans. Supported by LASA, we achieve state-of-the-art in real-world object reconstruction and 3D object detection.}
    \label{fig:teaser}
\end{center}%
}]
\begin{abstract}
Instance shape reconstruction from a 3D scene involves recovering the full geometries of multiple objects at the semantic instance level. Many methods leverage data-driven learning due to the intricacies of scene complexity and significant indoor occlusions. Training these methods often requires a large-scale, high-quality dataset with aligned and paired shape annotations with real-world scans. Existing datasets are either synthetic or misaligned, restricting the performance of data-driven methods on real data. To this end, we introduce LASA, a Large-scale Aligned Shape Annotation Dataset comprising 10,412 high-quality CAD annotations aligned with 920 real-world scene scans from ArkitScenes, created manually by professional artists. On this top, we propose a novel Diffusion-based Cross-Modal Shape Reconstruction (DisCo) method. It is empowered by a hybrid feature aggregation design to fuse multi-modal inputs and recover high-fidelity object geometries (see Fig.~\ref{fig:teaser}). Besides, we present an Occupancy-Guided 3D Object Detection (OccGOD) method and demonstrate that our shape annotations provide scene occupancy clues that can further improve 3D object detection. Supported by LASA, extensive experiments show that our methods achieve state-of-the-art performance in both instance-level scene reconstruction and 3D object detection tasks.

%extensive experiments, we demonstrate significant improvements in real-world instance-level shape reconstruction for DisCo and 3D Object detection for OccGOD supported by LASA dataset. Our codes and dataset will be publicly available upon acceptance. %It is time to embrace LASA!
\end{abstract}    
\vspace{-0.5cm}
% \begin{figure*}[ht]
%     \centering
%     \includegraphics[width=\linewidth]{fig/teaser.png}
%     \caption{Enter Caption}
%         \label{fig:teaser}
% \end{figure*}
\section{Introduction}
The widespread use of hand-held RGB-D sensors has facilitated the effortless acquisition of indoor scene scans. However, these scans often suffer from noises and incompleteness due to limitations in sensor accuracy, the complexity of indoor environments, and occlusion among objects. This further limits its applications in scenarios such as VR/AR and 3D industry where a complete and high-quality reconstruction is desired. This shortage absorbs great attention in 3D vision and graphics community, particularly in the realm of indoor instance-level scene reconstruction.  In this task, the objective is to reconstruct the shapes of observed objects based on 3D scans or images captured by sensors. Many advances~\cite{chibane2020implicit,yan2022shapeformer,xie2020pix2vox++,peng2020convolutional,zhang20233dshape2vecset} have been seen by leveraging the power of deep learning methods for this task. They are data-driven and demand a substantial number of paired scene scans and objects' CAD ground truths. 

Existing data-driven methods rely on two types of datasets. \cite{liu2022towards,xie2019pix2vox,xie2020pix2vox++,xu2019disn,wang2018pixel2mesh,gupta20233dgen,choy20163dr2n2,ranade2022ssdnerf,yan2022shapeformer,chibane2020implicit,arora2022multimodal,wu2020multimodal,cheng2023sdfusion} utilized synthetic datasets~\cite{chang2015shapenet,collins2022abo,3dfuture} by synthesizing images and scans that mimic real-world distributions.
Synthetic data provides CAD models perfectly aligned with input observations (images or scans) though. Models trained on it are vulnerable to domain gaps from the real world~\cite{wu2023scoda} that could impair the generalizability. Scan2CAD\cite{avetisyan2019scan2cad} attempts to bridge this gap by annotating objects' CAD in real-world scene scans. However, their CADs are retrieved from a synthetic database~\cite{chang2015shapenet} and are manually aligned to object scans. The retrieved shapes introduce misalignment issues. Many works~\cite{nie2021rfd,dong2023shape,tang2022point,tyszkiewicz2022raytran,gumeli2022roca,maninis2022vid2cad} using it as instance shape supervision are potentially biased with inferior alignment. In summary, The absence of a well-aligned real-world dataset barriers the further advancement of the community.

To surmount this challenge, we present a new dataset \textbf{LASA}, a \textbf{L}arge-scale \textbf{A}ligned \textbf{S}hape \textbf{A}nnotation dataset that contains 10,412 high-quality instance CAD annotations \textbf{meticulously crafted} by skilled artists. Each CAD shape is designed and placed to precisely align with objects' scans from 920 real-world scene scans in ArkitScene~\cite{baruch2021arkitscenes}. We deliberately annotate objects in ArkitScene instead of ScanNet\cite{dai2017scannet} because it provides scans obtained from both accurate Laser sensors and hand-held RGB-D sensors. The accurate Laser scan benefits high-quality and aligned manual annotations. Meanwhile, the less accurate scans from hand-held devices support research on reconstruction and completion using consumer-level devices.

A large-scale scan dataset with high-quality instance CAD annotations empowers many downstream applications. We first investigate how LASA benefits indoor instance-level scene reconstruction. Given an indoor scene, images and scene scans can be obtained through scanning. Typically, the initial step involves 3D object detection, after which the partial point clouds and multi-view images of each detected instance are acquired, serving as visual cues for subsequent shape reconstruction. Point cloud provides natural 3D information though, they are often noisy and incomplete. On the other hand, images present rich appearance clues but lack 3D constraints. Inspired by the complementary nature of these two modalities in describing object surfaces, we advocate utilizing both modalities as inputs to fuse their advantages. Recently, diffusion models have shown appealing performance in shape generation~\cite{zhang20233dshape2vecset,zheng2023las,gupta20233dgen,shue20233d,cheng2023sdfusion}. We advance it further for instance reconstruction and propose Diffusion-based Cross-modal Shape Reconstruction, namely \textbf{DisCo}. DisCo is a triplane diffusion model tailored to accommodate multi-modal inputs, towards robust real-world object reconstruction, where a hybrid feature aggregation layer is proposed to aggregate and align two input modalities effectively.
%, dedicated to local feature aggregation and alignment from the two input sources.
%It adopts volumetric representation to interact with pixel features and uses triplane representation for efficient point encoding.
Supported by the LASA dataset, extensive experiments show that our method achieves state-of-the-art reconstruction performance with real-world inputs.

Moreover, the LASA dataset has full annotations covering every instance within each of the 920 real-world scenes. This extensive coverage enables LASA to provide scene-level occupancy labels. We further explore how they can affect indoor scene 3D object detection. 3D Object detection \cite{qi2019votenet,rukhovich2022fcaf3d,wang2022cagroup3d} usually comprises a backbone and a detection head. We propose to integrate an occupancy prediction head for occupancy-guided 3D object detection, namely \textbf{OccGOD}. Our experiments demonstrate that incorporating occupancy prediction leads to substantial improvements in detection performance. In summary, our key contributions are four-fold:
%LISAScan can also benefit scene understanding tasks. Our well-aligned CAD models provide ground truth occupancy grids for 3D scenes. Occupancy grids represent objects using 3D cube collections with distinct geometric structures. This representation has shown effectiveness in boosting object detection in outdoor scenes~\cite{}. However, its usefulness for indoor scenes has been under-explored. By leveraging LISAScan, we are able to explore the impact of occupancy prediction on indoor object detection. Our experiments demonstrate that incorporating occupancy prediction leads to substantial improvements in detection performance.

\begin{itemize}[leftmargin=*]
\item We introduce a large-scale dataset, LASA. It contains 10,412 manually crafted, high-quality instance CAD annotations geometrically aligned with 920 real-world scene scans.
\item We propose DisCo, a novel diffusion-based method that leverages hybrid representation, effectively interacting with both input partial point cloud and multi-view images, achieving state-of-the-art reconstruction.
\item With LASA's scene-level annotation, we introduce occupancy-guided 3D object detection (OccGOD) with decent improvements. 
\item We strongly believe the large-scale dataset of well-aligned shape annotations can break the bottleneck of current research on 3D indoor scene understanding and reconstruction. 
\end{itemize}

\begin{table*}[ht]
\centering
\begin{tabular}{l | l | llll }
\hline
Dataset & Aligned & \#Scenes & \#CADs & Sensor Type & Annotation Method \\
\hline
Scan2CAD~\cite{avetisyan2019scan2cad} & - & 1,506 & 3,049 & RGB-D & Retrieval \\
CAD-Estate~\cite{maninis2023cadestate} & - & 19,512 & 12,024 & RGB & Retrieval \\
\hline
IKEA~\cite{girdhar2016learning} & \checkmark & null & 90 & RGB & Retrieval \\
Pix3D~\cite{sun2018pix3d} & \checkmark & null & 219 & RGB & Retrieval \\
ScanSalon~\cite{wu2023scoda} & \checkmark & 413 & 800 & RGB-D & Artist \\
Aria's Digital Twin~\cite{pan2023aria} & \checkmark & 2 & 370 & RGB-D & Artist \\
LASA (Ours) & \checkmark & 920 & 10,412 & RGB-D & Artist\\
\hline
\end{tabular}

\caption{Comparisons with existing 3D indoor datasets with instance shape annotations}
\label{tab:datasets}
\end{table*}
\section{Related Works}
\subsection{Indoor Instance Shape Dataset}
Recently, many advances have been seen in learning-based 3D object and scene reconstruction from images and videos. However, numerous challenges persist due to the limitations of existing datasets.

These learning-based approaches are usually trained either on existing synthetic datasets or real-world datasets. While synthetic datasets are demonstrated valuable for training models, they lack realism. Large synthetic object collections like ShapeNet\cite{chang2015shapenet} and ABO\cite{collins2022abo} provide diverse 3D models but lack environment context. Synthetic scene datasets such as 3D-Front\cite{fu20213dfront}, Replica\cite{straub2019replica}, and Structured3D\cite{Structured3D} possess complete synthetic environments with CAD annotations for objects. However, models trained on these datasets often struggle to generalize with real inputs due to substantial domain gaps\cite{wu2023scoda}.

Some works~\cite{avetisyan2019scan2cad,maninis2023cadestate,pan2023aria,wu2023scoda,sun2018pix3d,xiang2014pascal3d} annotate CAD models on real-world data to bridge the domain gaps. Scan2CAD~\cite{avetisyan2019scan2cad} and CAD-estate~\cite{maninis2023cadestate} have provided scanned object-CAD pairs by retrieving them from ShapeNet~\cite{chang2015shapenet} and are further manually aligned to the real-world scenes. However, these retrieved CAD models lack alignment with real objects, potentially biasing data-driven reconstruction methods towards inferior reconstruction. This motivates us to build a real-world dataset with well-aligned CAD annotations. Aria's Digital Twin~\cite{pan2023aria} and ScanSalon~\cite{wu2023scoda} provide such CADs though, the limited quantities restrict their application for data-hungry tasks. Datasets such as Pix3d~\cite{sun2018pix3d} and PASCAL3D+~\cite{xiang2014pascal3d} supply single-view images with aligned shapes but are limited in scene modalities lacking point cloud and multi-view images.

\subsection{3D Shape Reconstruction}
\textbf{Object-level Reconstruction}  Existing approaches leverage images or partial point clouds as input. Images based reconstruction methods accept either single-view~\cite{zhang2021holistic,nie2020total3dunderstanding,liu2022towards,xie2019pix2vox,zheng2023las,xu2019disn,pan2019deep,kurenkov2018deformnet,wang2018pixel2mesh,hane2017hierarchical,groueix2018papier,chen2019learning,mescheder2019occupancy,park2019deepsdf,gupta20233dgen} or sparse multi-view images~\cite{bautista2021generalization,xie2020pix2vox++,yang2022fvor,choy20163dr2n2,ranade2022ssdnerf} as inputs for shape reconstruction. They extract 2D image features and utilize them for shape reconstruction. Some~\cite{saito2019pifu,liu2022towards} leverage 2D pixel-aligned local features for reconstruction, demonstrating appealing performance for high-quality reconstruction. Other works~\cite{yan2022shapeformer,dai2017shape,litany2018deformable,rock2015completing,chibane2020implicit,wu2020multimodal,arora2022multimodal,zhang20233dshape2vecset} conduct shape reconstruction from noisy, incomplete partial point clouds. Both paradigms have their advantages, with image-based methods perceiving rich 2D appearance features, while point-based methods process native 3D signals. \cite{hou2020revealnet} proposed to combine both inputs to fuse their advantages. Our proposed approach also accepts both inputs, focusing on effectively aggregating and aligning local features from both input sources. Since diffusion model~\cite{cheng2023sdfusion,zhang20233dshape2vecset,zheng2023las,gupta20233dgen,shue20233d,wang2023rodin} have demonstrated strong capabilities in both shape generation and reconstruction, we opt for a diffusion model for robust and high-quality reconstruction.

\textbf{Instance-level Scene Reconstruction} Recent instance scene reconstruction methods~\cite{nie2021rfd,tang2022point,dong2023shape,maninis2022vid2cad,gumeli2022roca,tyszkiewicz2022raytran,hou2020revealnet} follow a detect-then-reconstruct pipeline from a single scan. First, these methods use a 3D object detector to localize objects in either scene scans or videos. Subsequently, the detected 3D objects are fed into a 3D shape reconstruction module individually to obtain object shapes. Finally, the reconstructed 3D object shapes are placed back at their original locations to obtain a reconstruction of the full scene. These methods commonly utilize the Scan2CAD \cite{avetisyan2019scan2cad} dataset for training, where the inferior alignment limits their reconstruction quality. In contrast, the CAD models in our LASA dataset are manually crafted by artists to guarantee alignment, which we hope can lay a foundation for future research in this area.\\
\section{LASA Dataset}

\begin{figure}
    \centering
    \includegraphics[width=1\linewidth]{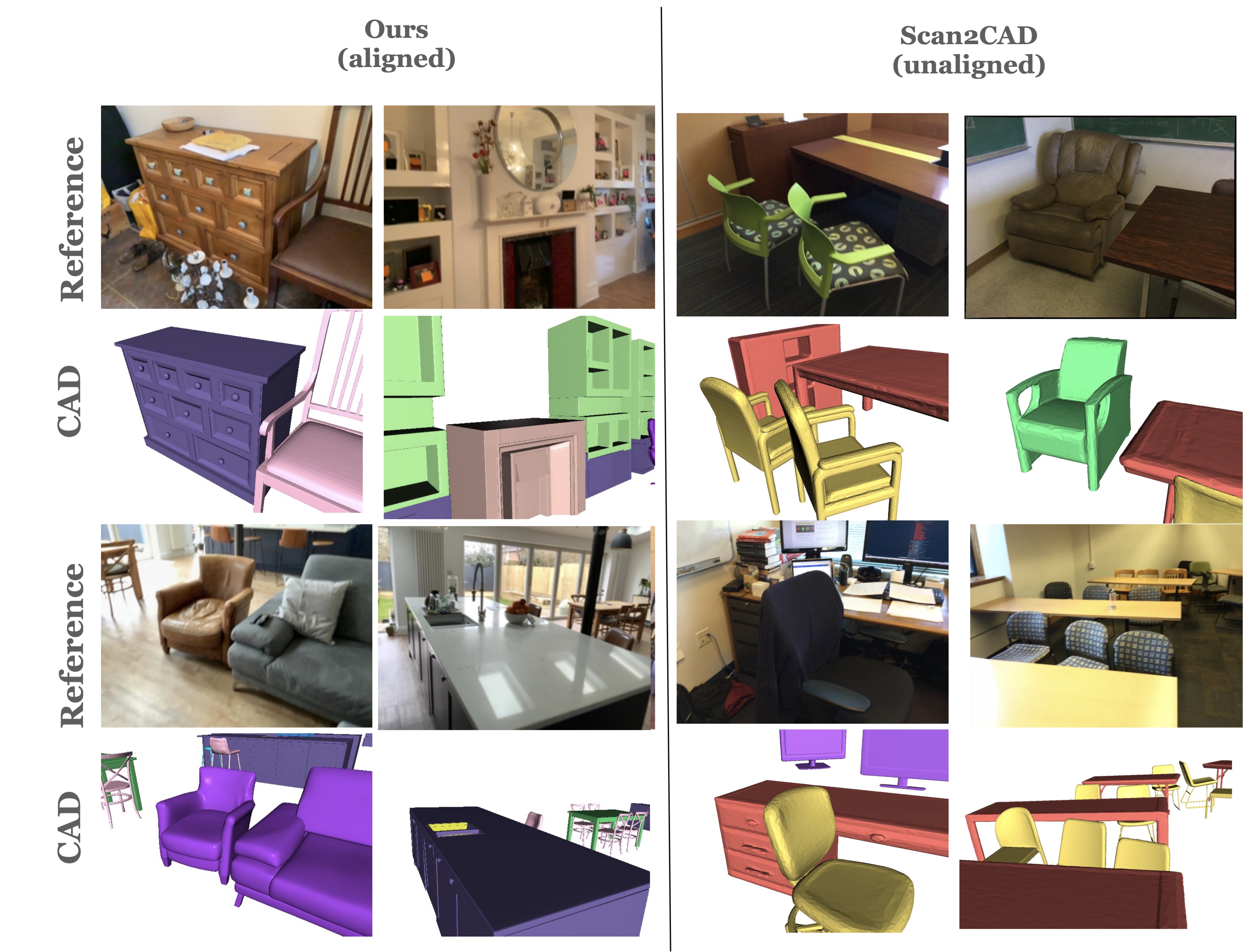}
    \caption{Visual comparison between aligned and unaligned CAD annotations}
    \label{fig:alignment}
    
\end{figure}

LASA is a large-scale dataset that contains 10,412 unique CAD models covering 920 scenes across 17 categories. 
Rather than relying on a pre-existing CAD database to annotate scenes, LASA engages professional artists to manually create aligned CAD models with 3D scans. Our annotations provide precise and consistent (as shown in Fig.~\ref{fig:alignment}) training data for data-driven reconstruction algorithms.
% The advantages of aligned CAD annotations are multifold. Firstly, it provides precise training and testing data for instance-level scene reconstruction. Secondly, annotations are not limited to what already exists in CAD libraries, which enables accurate annotations on all objects. In addition, specifically tailored CAD models show greater realism and diversity than the ones retrieved from a pre-built CAD libraries.

\subsection{Data Annotation}
LASA is built upon the 3D laser scans from ArkitScenes since it is accurate with high resolution, which is critical for high-fidelity CAD annotations. \\
\textbf{Data Preprocessing}. Each laser scan from ArkitScenes is exceptionally dense with 1GB+ data storage. To improve annotation efficiency, we downsample the scans to a 4mm density.
Since ArkitScenes does not publicly provide alignment transformations between the laser scans and the RGB-D scans, we utilize a coarse-to-fine registration method to calculate the transformation matrix (see our supplemental for details).
With the transformation matrix, we align the laser data with the RGB-D sequence coordinates from ArkitScenes. We then use ArkitScenes' 3D bounding box annotations to partition the aligned laser point clouds for each single object. These segmented point clouds are transformed into the canonical space. Furthermore, for each object, we select 2-5 frames from RGB-D scans that maximize the 2D projection area of its 3D bounding box. These selected frames serve as references for annotation. \\
% Initially, we employ Pointsect\yinyu{citation} to generate posed images from the comprehensive point cloud and register them with the RGB-D sequence using SfM\yinyu{citation}, resulting in an initial global transformation matrix. Subsequently, we employ a GICP\cite{} method to further refine the transformation matrix. 
\textbf{Shape Annotation.} The CAD annotation process involves a team of 35 artists working over 4 months. With preprocessed point clouds and reference images, each artist spent approximately 69 minutes designing a single model. Each model is annotated with Autodesk Maya or Cinema 4D.\\
\textbf{Shape Verification.}
We involve a shape verification procedure for annotation quality control. This procedure has both algorithmic validation and manual reviews to thoroughly evaluate CAD model's accuracy against ground truth scans and images. Our multi-step verification process includes

%Ensuring precision in our CAD annotations required rigorous alignment verification through manual reviews and computational validation. We implemented a robust quality assurance pipeline to thoroughly evaluate CAD model accuracy against ground truth scans and images. Our multi-step verification process included:
\begin{itemize}
\item Senior Review: After initial annotations, 6 senior designers reviewed every CAD model to verify quality, accuracy, and reliability by manually cross-checking against the 3D scans. Any models that failed to meet the standards were flagged for rework.
\item Geometry Alignment: We matched the CAD models to aligned laser scans and calculated the unidirectional Chamfer distance between them. This quantified the raw geometric alignment error for the CAD surface compared to the ground truth scan.
\item View Alignment: We also verified alignment in the pixel level, where we rendered 112,639 images across all scenes by positioning the CAD models in the view frustum of RGB-D sensors. Crowd workers performed a manual inspection to check if the rendered views overlay on the real images. They checked for inconsistencies along object edges and intricate details which would indicate misalignment. This pixel-level evaluation ensured precise alignment. Any rendering mismatches were fixed by re-annotating the CAD model.
\end{itemize}

\subsection{Dataset Statistics}
Tab.~\ref{tab:datasets} shows the statistics of LASA compared to the existing datasets. LASA contains a comparable number of unique CAD models to CAD-Estate~\cite{maninis2023cadestate}, which was previously the largest scene CAD dataset. 
However, LASA provides better CAD annotation quality against their retrieved CAD models. 
Additionally, LASA demonstrates greater shape diversity than Scan2CAD~\cite{avetisyan2019scan2cad}, with over 3 times as many unique CAD models.\\
Among all aligned datasets, LASA stands out with a total of 10,412 CADs. This is 13 times more than ScanSalon~\cite{wu2023scoda} and 28 times more than Aria's Digital Twin dataset\cite{pan2023aria}. Unlike IKEA~\cite{girdhar2016learning} and Pix3D~\cite{sun2018pix3d} which are annotated on single-view RGB images, LASA captures full RGB-D sequences. This enables a wider range of downstream applications compared to static image datasets.\\

\begin{figure*}[ht]
    \centering
    \includegraphics[width=\linewidth]{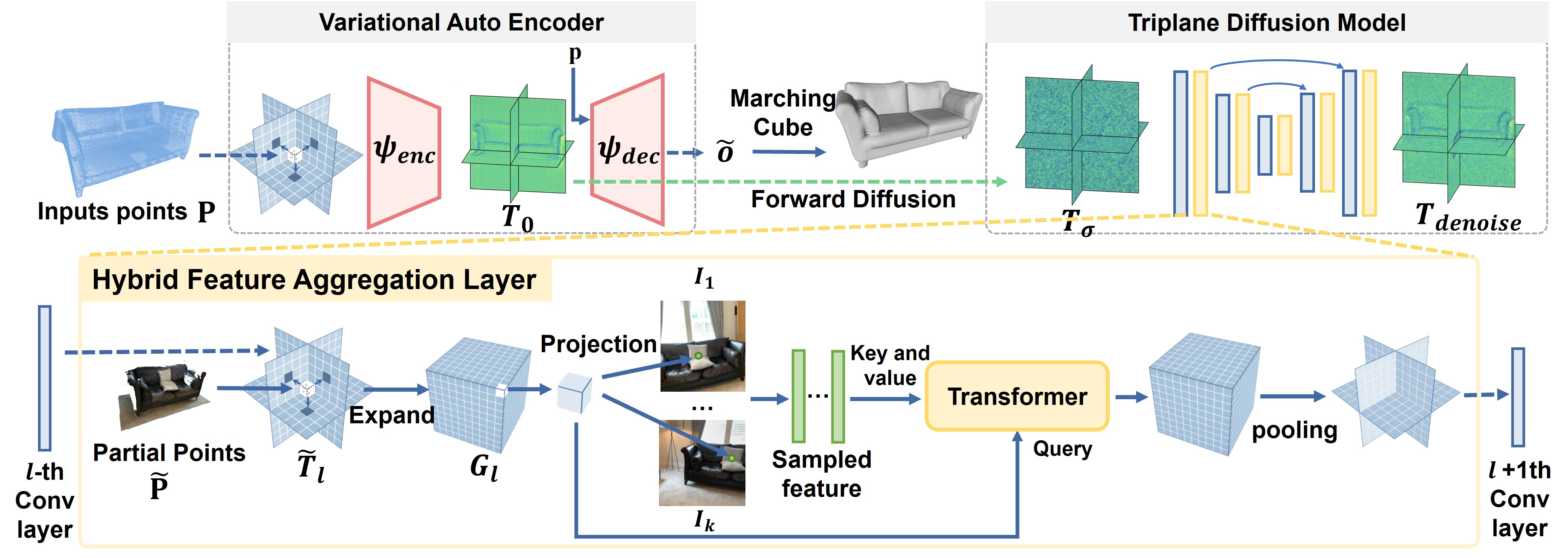}
    \caption{Pipeline of our DisCo. Firstly, a triplane VAE model is trained to encode shape into triplane latent space (top-left). Subsequently, a triplane diffusion model is trained in this latent space for conditional shape reconstruction (top-right). A novel Hybrid Feature Aggregation Layer is proposed to effectively aggregate and align local features in both partial points cloud and multi-view images (bottom).}
    \label{fig:triplaneDiffusion}
\end{figure*}

\section{Instance-level Scene reconstruction }
We propose a Diffusion-based Cross-modal Shape Reconstruction method (DisCo). DisCo is a diffusion-based model to pursue high-fidelity 3D shape reconstruction from partial point clouds and multi-view images. Various representation including latent sets~\cite{zhang20233dshape2vecset}, volumetric grids~\cite{cheng2023sdfusion,zheng2023las}, and triplanes~\cite{gupta20233dgen,wang2023rodin,ranade2022ssdnerf} are popular choice for diffusion model. We opt for the triplane since its efficiency enables higher output resolution compared to volumetric grids; while reserving 3D structure compared to latent sets. We employ latent triplane diffusion, where a triplane variational auto-encoder (VAE) first encodes shape into triplane latent space (in Sec.~\ref{sec:VAE}). Subsequently, a triplane diffusion model operates on this latent space for 3D shape reconstruction conditioned on both partial points and multi-view images (in Sec.~\ref{sec:diffusion}). The overall pipeline of DisCo is shown in the upper part of Fig.~\ref{fig:triplaneDiffusion}.

\begin{comment}
Reconstruction from individual modalities (e.g., point clouds~\cite{yan2022shapeformer,chibane2020implicit,zhang20233dshape2vecset} or multi-view images\cite{xie2020pix2vox++,bautista2021generalization,yang2022fvor,choy20163dr2n2,ranade2022ssdnerf}) have been well explored. However, relying solely on one modality often results in incomplete information, leading to ambiguous reconstruction. While partial points provide native 3d information, they are often partial and noisy. Conversely, multi-view images provide rich 2D information, but without 3D information. Their information are complementary. \yinyu{Sentences above already discussed in Introduction?}   
\end{comment}

It has been demonstrated that using multiple input modalities (images and scans) presents complementary benefits for semantic scene completion~\cite{hou2020revealnet}. In DisCo, we also fuse image and point features and introduce \textbf{Hybrid Feature Aggregation Layer} (in Sec.~\ref{sec:hybrid}). In this layer, we utilize a hybrid representation combining a triplane and a volumetric grid. This combination facilitates efficient local feature aggregation and feature alignment from both partial point clouds and multi-view images.
%to achieve high-fidelity reconstruction. 

\begin{comment}
Local features \cite{chibane2020implicit,peng2020convolutional,saito2019pifu} have been proven effective in achieving high-fidelity reconstruction. In the proposed triplane diffusion model, It is crucial to effectively utilize and align local features from both modalities to achieve high-fidelity reconstruction.\\
To this purpose, we introduce \textbf{Hybrid Feature Aggregation Layer} (in session \ref{sec:hybrid}). In this layer, we utilize hybrid representation combining triplane and volumetric grid. This combination facilitates efficient local feature aggregation and alignment from both partial point clouds and multi-view images, further enhancing the overall reconstruction accuracy. 
\end{comment}

\subsection{Triplane Variational Auto-Encoder}\label{sec:VAE}
To conduct a latent triplane diffusion, the first step is to learn an encoder capable of encoding shapes in a triplane latent space. The Triplane VAE~\cite{gupta20233dgen} comprises an encoder $\psi_{enc}$ and a decoder $\psi_{dec}$. The encoder processes inputs and encodes them into latent space, while the decoder recovers shape from the latent representation. Surface point cloud $\mathbf{P} \in \mathbb{R}^{K\times3}$ sampled from a ground truth shape, serves as inputs to $\psi_{enc}$. $K=20,000$ in our experiment. These points are projected onto a triplane and subsequently processed by a PointNet to form a triplane feature map. Followed by a UNet~\cite{ronneberger2015unet}, the encoder $\psi_{enc}$ outputs a normal distribution $\mathcal{N}(\mu,\sigma^2)$, where $\mu,\sigma \in \mathbb{R}^{H\times W \times 3 \times C}$. A latent triplane $T_0 \in \mathbb{R}^{H\times W \times 3 \times C}$ can be sampled from this distribution. The above process can be summarized as:
\begin{equation}
    T_0 \sim \mathcal{N}(\mu,\sigma^2)\quad\quad \mu,\sigma =\psi_{enc}(\mathbf{P})
\end{equation}
The occupancy field is chosen as the shape representation. The decoder $\psi_{dec}$ takes a point $p$ in space and the triplane latent $T_0$ as inputs, yielding the occupancy of this point. Specifically, A UNet model first refines $T_0$ and outputs a triplane feature map. Then, point $p$ is projected onto this triplane feature map, where features are sampled using bi-linear interpolation. These sampled features are fed into MLPs which outputs the occupancy prediction $\Tilde{o}$. The training of the VAE model is supervised by reconstruction loss and KL divergence loss as $\mathcal{L}_{vae}=\|\Tilde{o}-o_{gt}\|_2^2+\lambda_{kl}\mathcal{L}_{kl}$. In our experiment, $\lambda_{kl}=0.025$. As in \cite{wang2023rodin}, the network layers in both $\psi_{enc}$ and $\psi_{dec}$ adopt 3d aware convolution for triplane processing.

The reconstruction is conducted in canonical object space. During training, augmentation such as random shifting, rotating and scaling are applied, so that it will be more robust to inaccurate objects' pose during inference.
\subsection{Triplane Diffusion for reconstruction}\label{sec:diffusion}
Our approach employs latent triplane diffusion to reconstruct shapes based on partial point clouds and multi-view images as conditions. Specifically, a 2D UNet model serves as a denoise function $D(\cdot)$. This model comprises cascades of residual convolutional block and Hybrid Feature Aggregation Layer with 3d aware convolution. we introduce a novel Hybrid Feature Aggregation Layer, designed to foster effective interaction between local features derived from partial point clouds and multi-view images. We use continuous diffusion steps as in \cite{karras2022elucidating}. The forward diffusion process during training is defined as $T_{\sigma}=T_{0}+n$, where $n\sim \mathcal{N}(0,\sigma^2 \mathbf{I})$, $\sigma$ indicates the diffusion steps, and also the deviation of the noise added. The denoise UNet takes a noisy triplane latent $T_{\sigma}$ as input, and outputs a denoised triplane latent $T_{denoise}$. We follow the diffusion formulation in EDM \cite{karras2022elucidating}, the objective function during training is:
\begin{equation}
    \mathbb{E}_{\sigma,n,T_{0}}\lambda_\sigma\|D(T_{0}+n,\sigma,c)-T_{0}\|_2^2 \quad n\sim \mathcal{N}(0,\sigma^2 \mathbf{I})
\end{equation}
Here $c$ denotes the conditional inputs. In our case, they are partial point clouds and multi-view images. $\lambda_\sigma$ denotes a loss normalization factor. Inference can be conducted through sampling from noise triplane as in EDM ~\cite{karras2022elucidating}.

During training our diffusion model, we implement a strategy to first pretrain on synthetic datasets, then fine-tune on LASA dataset. We leverage ShapeNet \cite{chang2015shapenet}, ABO \cite{collins2022abo}, and 3D-Future \cite{3dfuture} dataset to synthesize partial point cloud and render images by embedding CAD models in HM3D scene\cite{yang2022hm3d,ramakrishnan2021hm3d}. This strategy enhances both the robustness and the performance of our method, enabling more effective real-world object reconstruction.

\begin{table*}[h]
\caption{Quantitative comparison on shape reconstruction. Evaluation metrics are mIoU / Chamfer L2 / F-score respectively. Higher mIoU and F-score are better while lower Chamfer L2 is better. Chamfer L2 is scaled by 1,000. LAS doesn't have mIoU value since it uses a surface occupancy representation.\label{tab:compare}}
\centering
\resizebox{1.0\linewidth}{!}{
\begin{tabular}{c|c|c|c|c|c|c}
\toprule
Method & Chair & Sofa & Table & Cabinet & Bed & Shelf \\
\midrule
IFNet & 28.9 / 19.8 / 25.2 & 66.1 / 3.61 / 31.1 & 28.3 / 21.4 / 27.4 & 73.8 / 3.20 / 36.6 & \textbf{64.0} / 2.95 / 28.8 & 17.2 / 5.24 / 29.6\\ 
LAS-pts & - / 13.2 / 22.6 & - / 4.47 / 22.9 & - / 19.5 / 23.9 & - / 4.81 / 22.1 & - / 4.77 / 25.8 & - / 4.70 / 26.8\\
LAS-pts+imgs & - / 10.7 / 24.1 & - / 3.76 / 24.1 & - / 15.7 / 24.6 & - / 4.56 / 24.4 & - / 4.58 / 26.5 & - / 4.35 / 27.2 \\
3DShape2VecSec & 32.5 / 6.43 / 25.6 & 66.7 / 3.39 / 27.6 & 30.9 / 12.9 / 25.3 & 68.1 / 4.09 / 23.7 & 62.3 / 3.69 / 30.5 & 22.3 / 3.79 / 32.0\\
Ours-pts & 33.4 / 5.68 / 26.6 & 68.8 / 3.21 / 29.0 & 38.0 / 10.5 / 32.2 & 72.3 / 3.63 / 36.5 & 54.0 / 2.97 / 34.1 & 23.2 / 3.68 / 36.5\\
Ours-pts+imgs & \textbf{38.6} / \textbf{3.57}/ \textbf{31.0} & \textbf{70.7}/ \textbf{2.88}/ \textbf{31.6} & \textbf{41.5} / \textbf{6.52} / \textbf{36.1} & \textbf{75.1} / \textbf{3.10} / \textbf{37.0} & 62.5 / \textbf{2.62} / \textbf{35.4} & \textbf{24.5} / \textbf{3.45} / \textbf{37.5} \\
\bottomrule
\end{tabular}}
\end{table*}

\subsection{Hybrid Feature Aggregation Layer}\label{sec:hybrid}
The objective of the Hybrid Feature Aggregation Layer is to aggregate and align local features from two input modalities to the triplane space. Points-triplane interaction is implemented by projecting points onto the triplane. However, such interaction becomes challenging when fusing images and triplanes. An alternative way is to expand the triplane into a volumetric grid and project the grids to images. This motivates us to propose this Hybrid Feature Aggregation Layer, introducing a hybrid representation that facilitates both points-triplane and images-triplane interactions.\\ 
As in the bottom part of Fig.~\ref{fig:triplaneDiffusion}, inputs of this layer are partial point cloud $\Tilde{P}$, k posed images $I_i, i\in{1,2...k}$, and triplane feature $\Tilde{T}_l$ from the l-th convolutional layer. It begins with the aggregation of local features from partial points, achieved by projecting them onto the triplane and incorporating a PointNet layer. Image features map is first extracted using pretrained Vision Transformer. Then, the triplane expands into a volumetric grids $G_l \in \mathbb{R}^{H\times\ W \times L\times C}$. The volumetric grids are then projected to k images using their camera poses, and image local features are sampled using bi-linear interpolation. 

To fuse features from multiple images, we employ a transformer. Specifically, the voxel feature serves as the query, while the sampled image feature serves as the key and value. The transformer outputs a volumetric grid attended with local image features from multi-view images. Finally, the volumetric grid is flattened back to a triplane by pooling, producing an output triplane feature map. This process ensures the effective aggregation and alignment of local features from diverse input modalities.

\section{Occupancy-guided 3D Object Detection}
3D object detection takes scene scans as input and parses the scene objects into 3D bounding boxes. 
In real-world scenarios, object scans are often incomplete and sparse due to occlusion, inaccurate sensors, and limited views during capture, making objects hard to recognize. To address it, we propose an Occupancy-Guided 3D Object Detection (OccGOD) approach that utilizes shape completeness prior for better scene understanding. 
We generate scene-level occupancy ground truth from LASA's fully-covered annotations. Specifically, scenes are partitioned into numerous 384x384x96 voxel grids, with a resolution of 4cm. All CAD models are then placed back into the scene, on which the surface points are densely sampled at 2cm intervals. Subsequently, we iterate through each point, marking their corresponding voxel to be occupied.  \\
Our methodology follows a simple `plug-and-play' manner.  It is compatible with any
detection method based on a 3D structured representation like volumetric grids~\cite{rukhovich2022fcaf3d,wang2022cagroup3d}. 
For ease of use, we built our OccGOD upon Cagroup3D~\cite{wang2022cagroup3d}. In addition to the original backbone and bounding box prediction head, we introduce an occupancy head and augment the bounding box prediction head with another two output parameters for orientation regression.  The occupancy heads takes backbone features as inputs, and output the occupancy of each voxel. More details of network design are in the supplemental. To further explore occupancy representations, we concatenate the features from the occupancy head with the backbone's features during ROI pooling for second-stage bounding box prediction. The occupancy head is supervised with a binary cross-entropy loss function with scene-level occupancy labels from LASA.

Our proposed OccGOD predicts a complete foreground structure to guide the bounding box detection. By leveraging the complete scene context, it achieves significant gains in detecting occluded and sparse objects compared to baseline methods.

\begin{figure*}
    \centering
    \includegraphics[width=\linewidth]{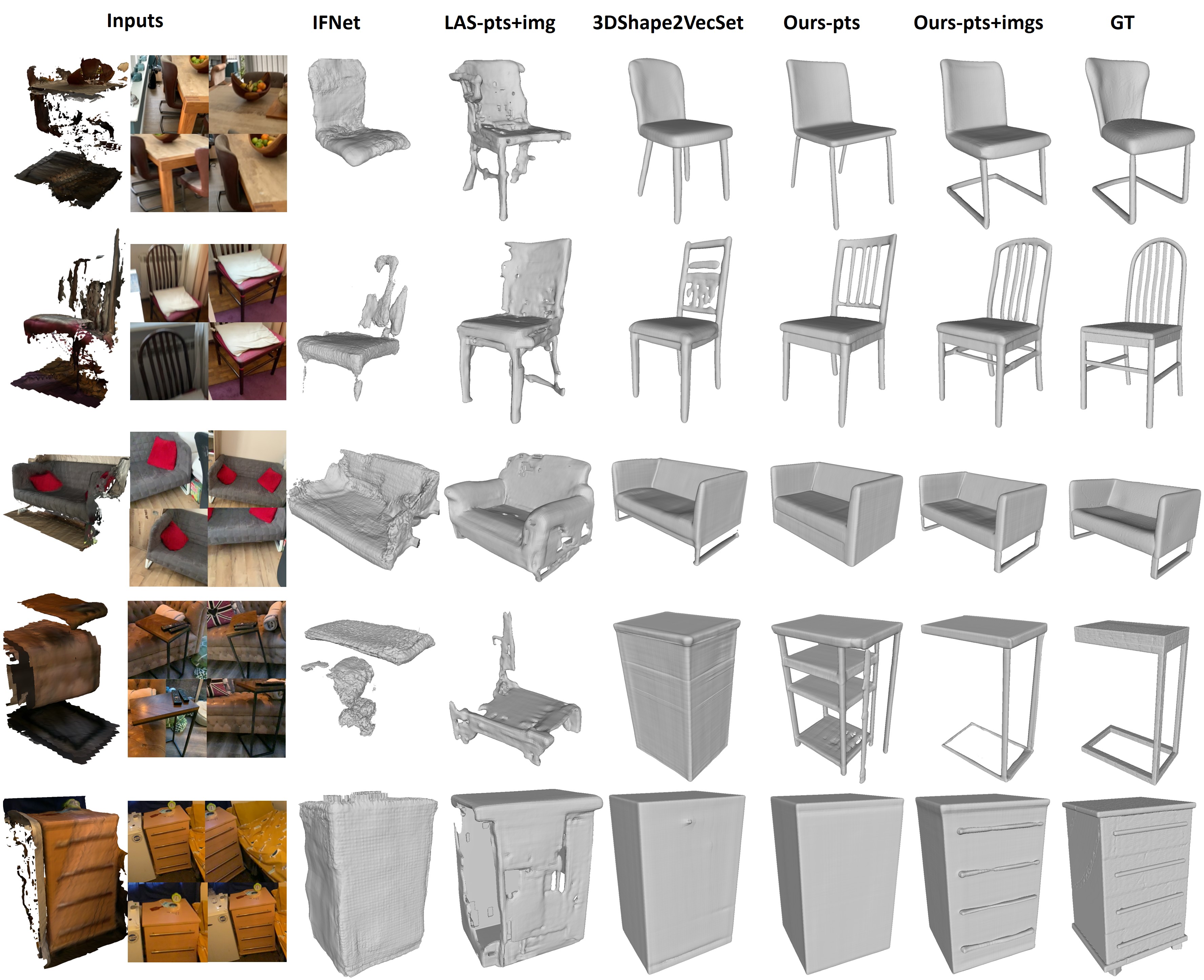}
    \caption{Qualitative comparison between our method and IFNet, 3DShape2Vecset, and LAS.}
    \label{fig:qualitative_comapre}
\end{figure*}

\section{Experiment}
\subsection{Experiment Setup}
For indoor object reconstruction, we employ mean Intersection over Union (mIoU), L2 chamfer distance, and 1\% F-score. We first normalize both results and ground-truth CAD into the interval from -0.5 to 0.5, and compute the above metrics between them. The experiment is conducted over 6 categories merged from 17 categories in LASA dataset.

For 3D object detection, we assess mean average precision (mAP) and mean average recall (mAR) with an IoU threshold at 0.5. Evaluations for both object reconstruction and 3D object detection are performed on LASA's test set.
\subsection{Evaluation on Indoor Object Reconstruction}
In this session, we compare our methods with existing baselines. We choose IFNet~\cite{chibane2020implicit} and two state-of-the-art diffusion-based methods, LAS Diffusion~\cite{zheng2023las} and 3DShape2VecSet~\cite{zhang20233dshape2vecset}, for comparison. We first pretrain all methods on synthetic dataset ~\cite{chang2015shapenet,collins2022abo,3dfuture}, then finetune them on LASA. IFNet and 3DShape2VecSet receive point clouds as inputs. For LAS Diffusion, we extend it into two versions: with point clouds as input (LAS-pts), and with both point clouds and images as input (LAS-pts+imgs). We compare them with our DisCo with two variants: 1) with partial point clouds only (Ours-pts); 2) with both modalities (Ours-pts+imgs). The quantitative and qualitative comparisons are shown in Tab.~\ref{tab:compare} and Fig.~\ref{fig:qualitative_comapre}. Our method achieves state-of-the-art performance both quantitatively and qualitatively.

\begin{table*}[ht]
\caption{Quantitative comparison between the state-of-the-art (CAGroup3D) and our OccGOD. Oriented bounding boxes are predicted. Evaluation metrics are mAP / mAR. Higher mAP / mAR indicates better performance.}
\resizebox{1.0\linewidth}{!}{
\begin{tabular}{c|c|c|c|c|c|c|c|c|c}
\hline
 $@IoU > 0.50$ & Chair & Table & Cabinet & Refrigerator & Shelf & Bed & Sink & Washer & Bathtub  \\
\hline
CaGroup3D &92.05 / 93.09 & 46.68 / 66.37 & 33.93 / 54.64 & \textbf{90.56} / 91.22 & 35.12 / 57.54 & 67.31 / 73.89 & 64.03 / 74.15& \textbf{87.77} / \textbf{89.39} & \textbf{27.99} / 45.97 \\
Our OccGOD & \textbf{92.57} / \textbf{93.25} & \textbf{49.17} / \textbf{66.76} & \textbf{35.05} / \textbf{55.90} & 90.45 / \textbf{91.89} & \textbf{38.15} / \textbf{58.19} & \textbf{70.78} / \textbf{78.33} & \textbf{70.30} / \textbf{78.74} & 86.27 / 88.64 & 26.56 / \textbf{46.77} \\
\hline
 $@IoU > 0.50$ & Toilet & Oven & Dishwasher & Fireplace & Stool & TV Monitor & Sofa & Stove & Overall \\
\hline
CAGroup3D & 50.94 / 68.99 & 78.93 / 81.28 & \textbf{92.19} / \textbf{92.86} & 28.21 / 45.19 & 70.69 / 80.78 & \textbf{1.16} / \textbf{6.63} & 46.32 / \textbf{65.51}& 33.32 / 41.73  &55.72 / 66.43 \\
Our OccGOD & \textbf{59.41} / \textbf{74.68} & \textbf{79.54} / \textbf{82.19} & 90.96 / 92.75 & \textbf{30.22} / \textbf{46.15} & \textbf{71.61} / \textbf{80.78}  & 0.73 / 5.19 & \textbf{47.65} / 64.98 & \textbf{35.64} / \textbf{41.73} & \textbf{57.36} / \textbf{67.47} \\
\hline
\end{tabular}
}
\label{tab:occgod}
\end{table*}           

\subsection{Ablation Study}
%We conduct several ablation studies to validate:

\textbf{Real-world Performance Boost using LASA}
We investigate the impact of our LASA dataset on real-world object reconstruction through three training setups: training from scratch on LASA (w/o pretrain), training on synthetic datasets only (w/o finetune), and pretraining on synthetic datasets followed by finetuning on LASA (full). The quantitative comparison is in Tab.~\ref{table:compare_strategies}. The table shows that finetuning on LASA significantly improves real-world reconstruction, with pretraining on synthetic data followed by finetuning on real strategy achieving the best performance.  

\begin{table}
\caption{Quantitative comparison on different training setups. The evaluation metrics are mIoU / chamfer L2 / F-score respectively.\label{table:compare_strategies}}
\centering
\resizebox{1.0\linewidth}{!}{
\begin{tabular}{c|c|c|c}
\toprule
Strategy & Chair & Sofa & Table \\
\midrule
w/o pretrain & 35.6 / 4.35 / 27.4 & 66.7 / 3.80 / 26.0 & 36.7 / 8.20 / 30.4\\
w/o finetune & 25.0 / 10.8 / 23.5 & 66.7 / 5.11 / 26.8 & 30.9 / 12.5 / 26.1\\
full & \textbf{38.6} / \textbf{3.57}/ \textbf{31.0} & \textbf{70.7}/ \textbf{2.88}/ \textbf{31.6} & \textbf{41.5} / \textbf{6.52} / \textbf{36.1} \\
\midrule
\midrule
 & Cabinet & Bed & Shelf \\
\midrule
w/o pretrain & 72.6 / 3.49 / 34.2 & \textbf{66.0} / \textbf{2.21} / 30.6 & 14.8 / 4.76 / 30.5\\
w/o finetune & 67.2 / 5.22 / 29.1 & 55.3 / 3.38 / 29.0 &  23.9 / 4.18 / 30.8\\
full & \textbf{75.1} / \textbf{3.10} / \textbf{37.0} & 62.5 / 2.62 / \textbf{35.4} & \textbf{24.5} / \textbf{3.45} / \textbf{37.5}\\
\bottomrule
\end{tabular}}
\end{table}

\textbf{Effectiveness of Hybrid Feature Aggregation Layer}
We verify the effectiveness of the Hybrid Feature Aggregation (HFA) layer in aggregating and aligning local features from both partial points and multi-view images. We compare it against not using the expanded grids to project onto the images. Specifically, the latter directly projects the triplane's pixels to the images. The quantitative comparison is shown in Tab.~\ref{table:hfa}. The decent improvement of the HFA layer verifies its powerfulness in fusing multi-modal features.

\begin{table}
\caption{Quantitative comparison between HFA layer and direct triplane projection. The evaluation metrics are mIoU / chamfer L2 / F-score respectively.\label{table:hfa}}
\centering
\resizebox{0.85\linewidth}{!}{
\begin{tabular}{c|c|c}
\toprule
Method & Chair & Sofa \\
\midrule
Triplane project & 38.1 / 3.95 / 30.6 & 70.0 / 3.03 / 30.8\\
HFA layer &  \textbf{38.6} / \textbf{3.57} / \textbf{31.0} & \textbf{70.7} / \textbf{2.88} / \textbf{31.6} \\
\bottomrule
\end{tabular}}
\end{table}

\textbf{Robustness to inaccurate detection}
We further investigate how inaccurate detection results could affect the reconstruction. An experiment is conducted by randomly rotating between -10 and 10 degrees, scaling between 0.8 and 1.1, and shifting the center between -10\% and 10\% for each object. The quantitative results are shown in Tab.~\ref{table:detection_error}. We observe that, with considerable disturbances on object poses, our method achieves robust accuracy.

\begin{table}
\caption{Quantitative comparison between reconstruction using GT 3D bounding boxes and noisy bounding boxes. Evaluation metrics are mIoU / chamfer L2 / F-score respectively.\label{table:detection_error}}
\centering
\resizebox{0.99\linewidth}{!}{
\begin{tabular}{c|c|c|c}
\toprule
Detection type & Chair & Sofa & Table \\
\midrule
noisy bbox & 38.7 / 3.77 / 31.3 & 70.2 / 2.90 / 31.8 & 40.6 / 7.42 / 34.4\\
GT bbox & 38.6 / 3.57 / 31.0 & 70.7/ 2.88/ 31.6 & 41.5 / 6.52 / 36.1 \\
\midrule
\midrule
 & Cabinet & Bed & Shelf \\
\midrule
noisy bbox & 73.1 / 3.70 / 34.4 & 65.5 / 2.64 / 36.2 &  24.1 / 3.85 / 36.0\\
GT bbox & 75.1 / 3.10 / 37.0 & 62.5 / 2.62 / 35.4 & 24.5 / 3.45 / 37.5\\
\bottomrule
\end{tabular}}
\end{table}      
\vspace{+0.5cm}
\textbf{Effectiveness of scene-level occupancy to OccGOD}
Tab.~\ref{tab:occgod} compares the baseline (Cagrounp3D) and our proposed OccGOD. Cagrounp3D achieves an mAP of 55.72 and an mAR of 66.43 with an IOU threshold of 0.5. Our OccGOD enhances the baseline, improving mAP by 1.64 and mAR by 1.04. Notable increases occurred for larger furniture like tables (+2.42 in AP), toilets (+8.47 in AP and +5.69 in AR), shelves (+3.03 in AP), beds (+3.47 in AP and 4.44 in AR), and sinks (+6.27 in AP and 4.59 in AR).

\subsection{More results}
\begin{figure}
    \centering
    \includegraphics[width=\linewidth]{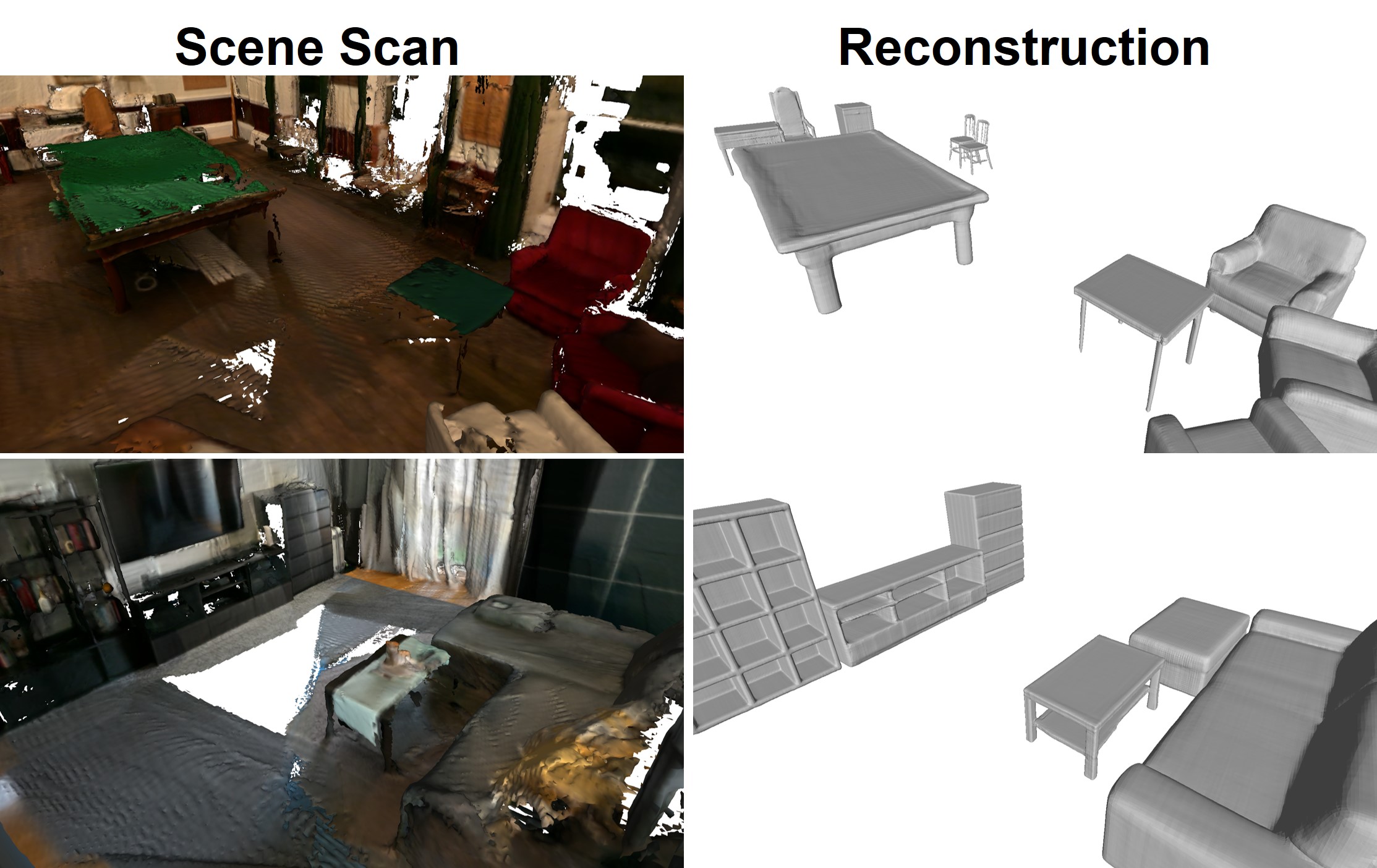}
    \caption{DisCo's instance-level scene reconstruction results. Both objects' partial point cloud and multi-view images are used as inputs to these results.}
    \label{fig:more_results}
\end{figure}

We show more DisCo's results on instance-level scene reconstruction as in Fig.~\ref{fig:more_results}. Noisy objects' poses are used to generate these results. DisCo supported by LASA is capable of producing high-quality and robust reconstruction results for real-world scenes.

\section{Conclusion}
We have introduced a new dataset LASA, a Large-scale Aligned Shape Annotation Dataset. In this work, we have illustrated the substantial benefits LASA brings to the community, particularly in the realms of indoor instance-level scene reconstruction and 3D object detection. Empowered by LASA, we propose a novel Diffusion-based Cross-Modal Shape Reconstruction approach, namely DisCo, and an Occupancy-guided 3D Object Detection method, namely OccGOD. In DisCo, we design a novel Hybrid Feature Aggregation Layer to effectively fuse and align local features from two input modalities - partial point clouds and multi-view images. In OccGOD, we leverage the scene-level occupancy labels provided by LASA, to enhance 3D object detection by learning object completeness priors. Extensive experiments demonstrated that, with the support of LASA, both methods achieve state-of-the-art performance in real-world scenarios.

We firmly believe that the large-scale and well-aligned features of LASA present better annotation quality and quantity, laying a foundation for many 3D downstream applications, including 3D understanding and reconstruction.% The contributions presented in this work pave the way for further advancements in the field.
{
    \small
    \bibliographystyle{ieeenat_fullname}
    \bibliography{main}
}


\end{document}